\documentclass[runningheads]{llncs}

\usepackage{eccv}

\usepackage{eccvabbrv}

\usepackage{graphicx}
\usepackage{booktabs}

\usepackage[accsupp]{axessibility}  

\usepackage[pagebackref,breaklinks,colorlinks,citecolor=eccvblue]{hyperref}

\usepackage{orcidlink}

\usepackage{adjustbox}
\usepackage{xcolor}
\usepackage{multirow}
\usepackage{hyperref}
\usepackage{makecell}
\usepackage{amsmath,amsfonts}
\usepackage{algorithmic}
\usepackage{algorithm}
\usepackage{array}
\usepackage{textcomp}
\usepackage{stfloats}
\usepackage{url}
\usepackage{verbatim}
\usepackage{graphicx}
\usepackage{cite}
\usepackage{caption}
\usepackage{bm}
\usepackage[bb=dsserif]{mathalpha}
\usepackage[T1]{fontenc}
\usepackage{aecompl}
\usepackage{booktabs}
\usepackage{wrapfig}

\usepackage{tikz}
\usetikzlibrary{calc}
\usetikzlibrary{shapes.geometric, arrows, decorations.pathreplacing}
\usetikzlibrary{backgrounds}
\definecolor{network-blue}{RGB}{165, 192, 221}
\definecolor{light-yellow}{RGB}{238, 233, 218}
\definecolor{yellow}{RGB}{230, 220, 218}
\definecolor{light-green}{RGB}{211, 235, 205}
\definecolor{light-red}{RGB}{242, 182, 160}
\definecolor{light-blue}{RGB}{124, 150, 171}
\definecolor{green}{RGB}{105,135,105}
\definecolor{red}{RGB}{231,97,97}
\definecolor{dark-red}{RGB}{181,23,0}
\definecolor{environment-c}{RGB}{254, 242, 244}
\definecolor{emb-c}{RGB}{0, 51, 124}
\definecolor{gt-c}{RGB}{155, 164, 181}
\definecolor{standard-gaussian}{RGB}{230, 226, 195}
\definecolor{prior-distribution}{RGB}{136, 164, 124}
\definecolor{skill-instance}{RGB}{28, 49, 94}
\begin{document}
\title{What Matters to Enhance Traffic Rule Compliance of Imitation Learning for End-to-End Autonomous Driving} 

\titlerunning{Imitation Learning with Penalties}

\author{Hongkuan Zhou\inst{2}\thanks{Equal contribution}\orcidlink{0000-0002-3665-9822} \and
Wei Cao\inst{2}\textsuperscript{$\ast$}\orcidlink{0009-0005-5163-6484}\and
Aifen Sui\inst{1}\textsuperscript{$\ast$}\orcidlink{0009-0006-2094-9178}\and
Zhenshan Bing\inst{2}\orcidlink{0000-0002-0896-2517}}

\authorrunning{H.~Zhou et al.}

\institute{Huawei Munich Research Center, Munich, Germany \\
\email{aifen.sui@huawei.com} \and
Technical University of Munich, Munich, Germany \\
\email{zhouh@in.tum.de}, \email{wei.cao@tum.de}, and \email{zhenshan.bing@tum.de}
}
\maketitle

\begin{abstract}
End-to-end autonomous driving, where the entire driving pipeline is replaced with a single neural network, has recently gained research attention because of its simpler structure and faster inference time. 
Despite this appealing approach largely reducing the complexity in the driving pipeline, it also leads to safety issues because the trained policy is not always compliant with the traffic rules. 
In this paper, we proposed P-CSG, a penalty-based imitation learning approach with contrastive-based cross semantics generation sensor fusion technologies to increase the overall performance of end-to-end autonomous driving. In this method, we introduce three penalties - red light, stop sign, and curvature speed penalty to make the agent more sensitive to traffic rules. The proposed cross semantics generation helps to align the shared information of different input modalities. We assessed our model's performance using the CARLA Leaderboard - Town 05 Long Benchmark and Longest6 Benchmark, achieving 8.5\% and 2.0\% driving score improvement compared to the baselines. Furthermore, we conducted robustness evaluations against adversarial attacks like FGSM and Dot attacks, revealing a substantial increase in robustness compared to other baseline models. 
\end{abstract}

\section{Introduction}\label{sec:introduction}
End-to-end autonomous driving \cite{hu2022st} integrates the perception and decision-making layers into one deep neural network.
The perception component extracts essential information about the surrounding environment. Despite some approaches \cite{DBLP:journals/corr/abs-2109-08473} \cite{DBLP:journals/corr/abs-2008-05930}, which leverage LiDAR sensor input and HD maps and demonstrate impressive performance, relying on HD maps is not a feasible option as it requires substantial resources to create and maintain. They may not be universally available for all areas and regions. 
Recent research has concentrated on multi-modality sensor techniques utilizing both LiDAR \cite{li2023lwsis,fu2024eliminating} and cameras \cite{sobh2018end, Chitta2022PAMI, shao2022interfuser, chen2022learning}. To manage the complexity of LiDAR input, various approaches have been developed, including point-based \cite{Xie_Xiang_Yu_Xu_Yang_Cai_He_2020}, voxel-based \cite{8794195, yoo20203d, zhu2022vpfnet, wang2021pointaugmenting}, and range-view-based approaches \cite{fadadu2022multi, chen2020mvlidarnet, sun2020fuseseg}.

More recently, with the popularity of attention mechanisms, many researchers are trying to use Transformer \cite{vaswani2017attention,xie2021segformer} to integrate multimodal information \cite{Chitta2022PAMI} \cite{Prakash2021CVPR} \cite{shao2022interfuser}. 
Despite its potential, the Transformer architecture on a large scale leads to a significant increase in both the training and inference time. In this paper, we shift our approach by extracting and aligning shared information from diverse modalities by contrastive learning, enabling the model to extract the global context within multi-modalities.

\begin{wrapfigure}{r}{0.5\textwidth}
    \centering
    \begin{adjustbox}{width=0.5\textwidth}
    \begin{tikzpicture}[node distance=2cm]
        \node[inner sep=0pt] (figure1){\includegraphics[width=8cm]{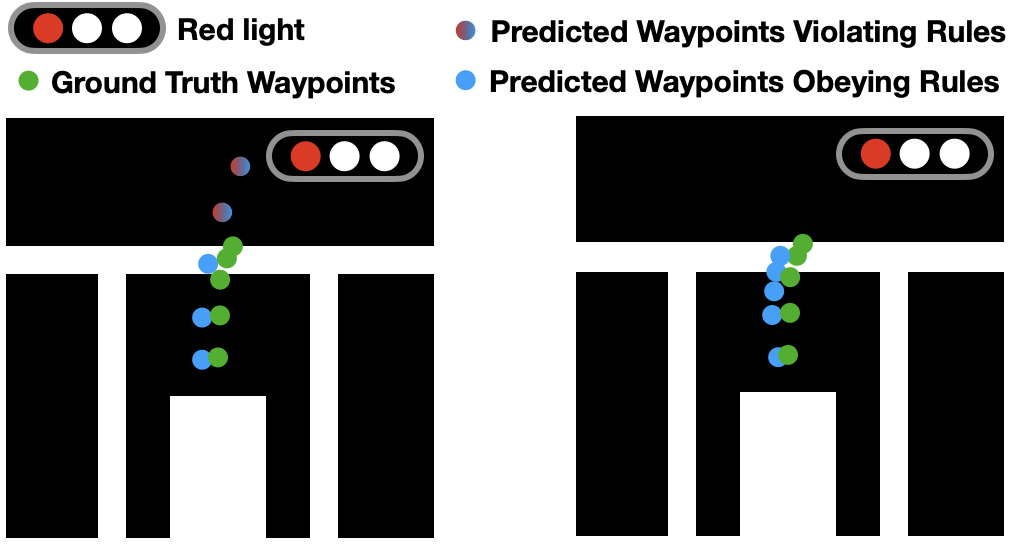}};
        \node[below of=figure1, xshift=-2.25cm, yshift=-0.5cm]{\small (a)};
        \node[below of=figure1, xshift=2.25cm, yshift=-0.5cm]{\small (b)};
    \end{tikzpicture}
    \end{adjustbox}
    \caption{(a) Common imitation learning approach could learn the wrong behavior which violates the traffic rule. (b) Our approach penalizes the predicted waypoints violating traffic rules to enhance the agent's traffic rule adherence.}
    \label{fig:examples}
    \vspace{-1em}
\end{wrapfigure}
Imitation learning \cite{Imitation_Learning} and reinforcement learning \cite{sutton1998introduction} are two main learning paradigms utilized by end-to-end autonomous driving systems for the decision-making part.
Reinforcement learning suffers from the low sample efficiency and the complexity of reward signs. It may also learn risky actions, leading to accidents and unsafe driving.
Due to these shortages, other researchers have turned to imitation learning approaches \cite{vaswani2017attention, Chitta2022PAMI, Prakash2021CVPR, shao2022interfuser}. However, these approaches still lack effective mechanisms to ensure the agent to comply the traffic rules. 
For one thing, the expert policy to generate demonstrations may still make mistakes when collecting the training data. 
Also, the metrics for autonomous driving (e.g., traffic rule violations) and the objectives of imitation learning (minimize the difference between the predicted and ground truth waypoints) remain disparate, indicating that a low loss in the learning objective does not guarantee optimal performance of the agent in the testing environments. 

To overcome the above-mentioned limitations, we propose \textbf{P}enalty-based Imitation Learning with contrastive-based \textbf{C}ross \textbf{S}emantics \textbf{G}eneration (\textbf{P-CSG}).
Our approaches penalize the behavior (higher loss) that violates the traffic rules during the imitation learning process to make the agent better adhere to these traffic rules (cf. Figure \ref{fig:examples}).
Meanwhile, we adopt a contrastive-based sensor fusion technology that aligns the shared information of multi-modalities so that the model can better understand the global context within multi-modalities. 

We evaluated our model on the CARLA Leaderboard - Town 05 Long Benchmark and Longest6 Benchmark. It achieves outstanding performance, especially in collision and traffic rule violations. We also analyze the performance and robustness of our autonomous driving model under two types of adversarial attacks: Fast Gradient Sign Method
(FGSM) Attack \cite{goodfellow2015explaining} and Dot attacks \cite{li2019adversarial}. The results show that our model outperforms the other baselines from the safety perspective. 

Our main contributions to this paper can be summarized as follows:
\begin{itemize}
    \item Our innovative multi-sensor fusion technique involves aligning shared information from various modalities, thus extracting the global context across diverse modalities.
    \item We proposed a penalty-based imitation learning approach that leverages constraint optimizations to make the imitation learning model more sensitive to traffic rule violations. 
    \item We evaluate our method's performance under adversarial attacks w.r.t. FSGM and Dot Attacks. The results show that our model adopts a more cautious strategy and has a more robust performance.
\end{itemize}

The rest of this paper is organized as follows: Section \ref{sec:related works} offers a detailed review of related literature. We present our methodology in Section \ref{sec:methodologies}. Section \ref{sec:experiments} provides the experiment results and discussions. Section \ref{sec:robustness study} conduct robustness study against adversarial attacks. We conclude our paper in Section \ref{sec:conclusion}.

\section{Related Works}
\label{sec:related works}
Imitation learning \cite{sobh2018end, Prakash2021CVPR, chen2022learning, Chitta2022PAMI, shao2022interfuser, DBLP:journals/corr/abs-1812-03079} are widely used in the field of autonomous driving \cite{SOS, SOS1, SOS2, Planning}. Many approaches use the CARLA simulator \cite{Dosovitskiy17} to collect training data and test the performance of the trained agent in the simulation environment. To improve the agent's overall performance, many approaches investigate the multi-sensor fusion technologies and safety mechanisms.

\subsection{Multi-sensor Fusion Technologies}
Multi-sensor fusion has received much research attention in the field of end-to-end autonomous driving. 
Recent works \cite{xiao2020multimodal, behl2020label, zhou2019does, natan2022end}, complementing camera images with depth and semantics has shown the potential to improve the overall driving performance. 
More research attention has been given to LiDAR and camera input fusion since they are complementary to each other in terms of representing the scene and are readily available in autonomous driving systems. 
Early works such as LateFusion \cite{sobh2018end} concatenate the feature embeddings from different modalities and use a Multi-Layer-Perception (MLP) network to weigh, select, and fuse the concatenated features. 
TransFuser \cite{Prakash2021CVPR} leverage Transformer-based approaches to achieve multi-modal global context. They apply cross-attention to the feature embeddings from different modalities in different layers. TransFuser+ \cite{Chitta2022PAMI}, as an extension of TransFuser, introduces auxiliary tasks: depth prediction and semantic segmentation are performed with image information, while HD map prediction and vehicle object detection are handled with bird's eye view (BEV) LiDAR information. These auxiliary tasks serve to illuminate the inner workings of the entire network. Similarly, InterFuser \cite{shao2022interfuser} also adopts the Transformer-based sensor fusion approach. 

Contrastive Learning, which is widely used in vision language models (VLMs) \cite{radford2021learning, li2022blip}, demonstrates its strong performance in multi-modal information handling. Some recent works \cite{liu2021contrastive, 9976192, li2022clmlf} focus on using contrastive learning for multi-modal information fusion. In this paper, we investigate contrastive learning for LiDAR and camera input fusion and propose a contrastive-based cross semantic generation sensor fusion method to extract the global context within the modalities of LiDAR and camera. 

\subsection{Safety Mechanism}
To bolster the safety of autonomous driving agents, researchers in \cite{Chitta2022PAMI} concentrate on refining the expertise of the expert agent. By filtering out incorrectly demonstrated actions executed by the expert policy, the agent avoids being misled and can more effectively acquire skills from the demonstrations. LAV \cite{chen2022learning} supervises waypoint outputs with additional data by making predictions for other nearby agents. InterFuser \cite{shao2022interfuser} developed a safety control module to regulate the agent's behaviors, preventing the agent from violating the traffic rules. However, this control module heavily depends on hand-designed heuristics and can vary in different systems. In our paper, we introduce the `penalty' concept to the IL framework, which incentivizes the trained agent to adopt safer driving behaviors in an end-to-end manner. 
\subsection{Combining IL and RL}
Methods such as DQfD\cite{hester2018deep}, DDPGfD\cite{vecerik2017leveraging}, and DAPG\cite{rajeswaran2017learning} have shown that integrating IL with RL can help address exploration challenges in domains with sparse rewards. Other offline RL approaches, such as TD3+BC \cite{fujimoto2021minimalist} and CQL \cite{kumar2020conservative}, combine RL and IL objectives to regularize Q-learning updates and prevent overestimation of out-of-distribution values. In the autonomous driving field, Lu et al.~\cite{https://doi.org/10.48550/arxiv.2212.11419} proposed behavior cloned soft actor-critic (BC-SAC), which incorporates a behavioral cloning objective into the SAC objective.

Our approach starts with an IL perspective by directly incorporating the concept of penalties and rewards from RL into IL. We model penalties as constraints and violating these constraints leads to higher loss. Q-function learning is not included in our approach. 
\section{Methodologies}
\label{sec:methodologies}
In this section, we propose a multi-sensor fusion approach and a penalty-based imitation learning method for end-to-end autonomous driving.


\subsection{Problem Setting}
We concentrate on point-to-point navigation in an urban setting where the goal is to complete a route with safe reactions to dynamic agents such as moving vehicles and pedestrians. Traffic rules should be followed, such as red lights and stop signs. We consider the problem of learning a goal-conditioned policy $\pi(\boldsymbol{a}|\boldsymbol{s}, \bm{g})$ that outputs action $\boldsymbol{a} \in \mathcal{A}$, conditioned on the current observed state $\boldsymbol{s} \in \mathcal{S}$ and a goal location $\bm{g}$ provided by GPS. The environment we used to train and test the performance of the trained agent is CARLA. The specific environment setting we used can be characterized by the following statements:
\begin{itemize}
    \item A state space $\mathcal{S} \subset \{(\mathbb{R}^{H_\textrm{c} \times W_\textrm{c} \times 3}, \mathbb{R}^{H_\textrm{L} \times W_\textrm{L} \times 2}, \mathbb{R}^4)\}$ which is the combination of the observation of camera with the shape of $H_\textrm{c} \times W_\textrm{c} \times 3$, bird's eye view (BEV) LiDAR pseudo image input with the shape of $H_\textrm{L} \times W_\textrm{L} \times 2$, and measurements of current throttle, brake, steer, and speed.
    \item A multidimensional action space $\mathcal{A} \subset \mathbb{R}^3$. The action space contains the parameters of steer, throttle, and brake to drive the agent to finish tasks. Instead of directly predicting the next action of throttle, steer, and brake, we estimate the future waypoints $\mathcal{W}$ of the ego-vehicle in BEV space, which is centered at the ego vehicle's current coordinate frame. Then, we use two fine-tuned PID controllers for lateral and longitudinal control to obtain steer, throttle, and brake values from the predicted waypoints.
    \item A goal space consists of high-level goal location provided as GPS coordinates. 
\end{itemize}

\subsection{Policy Learning}
In this paper, we use the imitation learning mechanism to learn the goal-conditioned policy $\pi(\bm{a}|\bm{s},\bm{g})$. As we discussed previously, the future waypoints instead of the actions are predicted. An expert policy is applied in the environment to collect a dataset $\mathcal{D}=\{(\bm{s}, \mathcal{W})\}^{N}$ with the size of $N$, which contains tuples of the observation $\bm{s}$ and a set of waypoints $\mathcal{W}$ in future time steps. Similar to previous works \cite{Chitta2022PAMI}, \cite{chen2019lbc}, we use $l_1$-norm based loss function because of its demonstrated robustness to outliers. For each input, the policy learning loss can be formalized as:
\begin{equation}
    \label{eq:policy learning loss}
    \mathcal{L}_{\textrm{pl}} = \sum_{t=1}^{T} ||\bm{\hat{w}_t} - \bm{w_t}||_1 \text{,}
\end{equation}
where $\bm{\hat{w}_t}$ is the $t$-th predicted waypoint and $\bm{w_t}$ is the \textit{t}-th ground truth waypoint produced by the expert policy.

\tikzstyle{emb} = [rectangle, minimum height=0.5cm, align=center, draw=black, fill=emb-c!60, font=\scriptsize\linespread{0.9}\selectfont]
\tikzstyle{module} = [rectangle, rounded corners, text centered, draw=black, minimum height=1cm, fill=network-blue!70]
\tikzstyle{dash-box} = [rectangle, dashed, thick, draw=emb-c!60]
\tikzstyle{arrow} = [thick,->,>=stealth]

\begin{figure*}[ht]
    \centering
    \resizebox{1.0\textwidth}{!}{
    \begin{tikzpicture}[node distance=2cm]
        \node (feature-extraction-module) [rectangle, rounded corners, fill=light-yellow!50, draw=light-yellow, very thick, minimum height=5.5cm, minimum width=8.5cm, label={[label distance=0cm, xshift=1cm, yshift=-0.5cm]130:\small Feature Extraction}] {};
        \node (cross-semantic-generation-module) [rectangle, rounded corners, fill=light-green!20, draw=light-green, very thick, minimum height=7cm, minimum width=4.5cm, below of = feature-extraction-module, xshift=2cm, yshift = -4.75cm] {};
        \node (cross-semantic-generation-module2) [rectangle, rounded corners, fill=light-green!20, draw=light-green, very thick, minimum height=2.25cm, minimum width=8.5cm, below of = feature-extraction-module, xshift=0cm, yshift = -7.125cm, label={[label distance=0cm, xshift=1cm, yshift=0.5cm]-135:\small Cross Semantic Generation}] {};
        \node (cross-semantic-generation-module3) [rectangle, rounded corners, fill=light-green!20, minimum height=0.5cm, minimum width=4.45cm, below of = feature-extraction-module, xshift=2cm, yshift = -6cm] {};
        \node (auxiliary-module) [rectangle, rounded corners, fill=light-red!10, draw=light-red, very thick, minimum height=4.25cm, minimum width=3.5cm, below of = feature-extraction-module, yshift = -3.375cm, xshift=-2.5cm, label={[label distance=0cm, xshift=1.75cm, yshift=0.125cm]-135:\small Auxiliary}] {};
        
        \node (penalty-based-imitation-learning-module) [rectangle, rounded corners, fill=light-red!10, draw=light-red, very thick, minimum height=13cm, minimum width=8.5cm, right of = feature-extraction-module, xshift = 7cm, yshift=-3.75cm, label={[label distance=0cm, xshift=5cm, yshift=1.7cm]135:\small Penalty-based Imitation Learning}] {};

        \node[inner sep=0pt, above of = feature-extraction-module, xshift=-1.5cm, yshift=-0.625cm] (rgb-image)
        {\includegraphics[width=4.5cm]{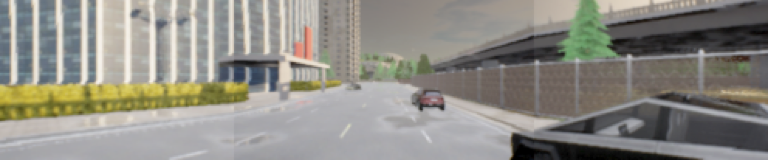}};
        
        \node[inner sep=0pt, right of = rgb-image, xshift=2.25cm, yshift=0.25cm] (lidar-image1)
        {\includegraphics[width=1.5cm]{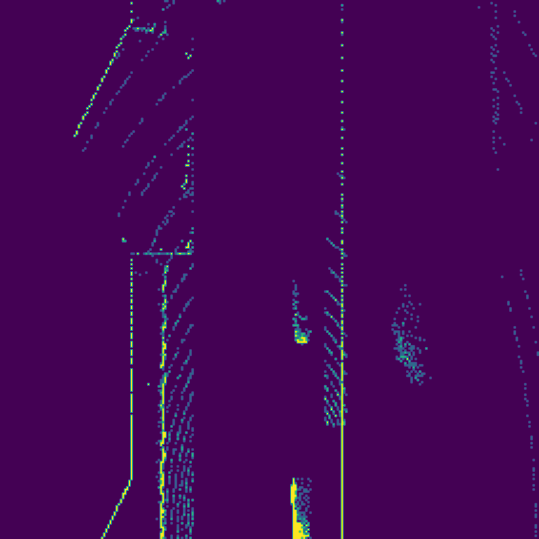}};
        
        \node[inner sep=0pt, right of = rgb-image, xshift=2.5cm, yshift=0cm] (lidar-image1)
        {\includegraphics[width=1.5cm]{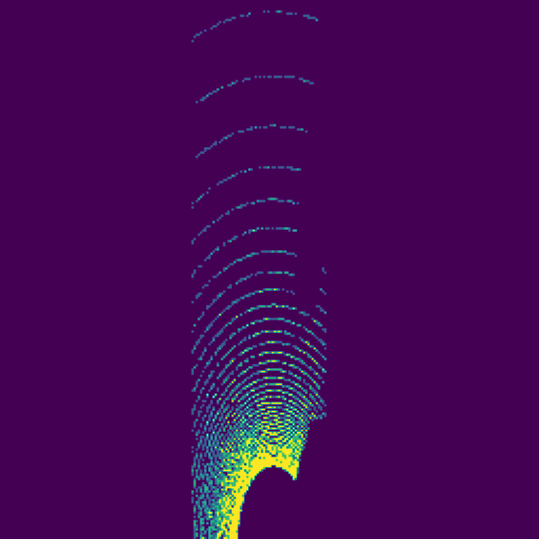}};
        \node (resnet50) [module,  below of=rgb-image, align=center, minimum width=4.5cm, yshift=0.25cm] {Resnet50};
        \node (resnet18) [module,  below of=lidar-image1, align=center, minimum width=1.75cm, yshift=0.25cm, xshift=-0.125cm] {Resnet18};
        \node (rgb-emb) [emb, below of=resnet50, yshift=0.5cm, minimum width=3cm, xshift=-0.5cm] {\textcolor{white}{512}};
        \node (lidar-emb) [emb, right of=rgb-emb, minimum width=3cm, xshift=2cm] {\textcolor{white}{512}};
        \node (dash-box1) [dash-box, right of=rgb-emb, xshift=0cm, yshift=0cm, minimum width=7.5cm, minimum height=1cm]{};

        \draw [arrow] (rgb-image.south) -- (resnet50.north);
        \draw [arrow,draw=gray] (lidar-image1.south -| resnet18.north) -- (resnet18.north);
        \draw [arrow] (resnet50.south) -- +(0, -0.25) -| (rgb-emb.north);
        \draw [arrow, draw=gray] (resnet18.south) -- +(0, -0.25) -| (lidar-emb.north);


        \node (decoder) [module,  above of=auxiliary-module, align=center, minimum width=2.5cm, yshift=-0.75cm] {Decoder};
        \node[inner sep=0pt, below of = decoder, xshift=-0.75cm, yshift=-0.125cm] (stop-sign)
        {\includegraphics[width=1.4cm]{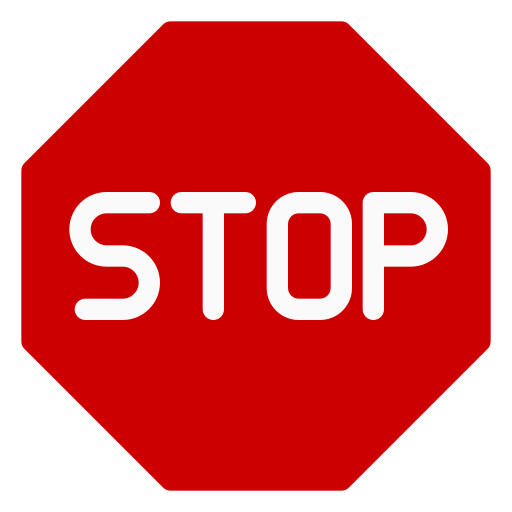}};
        \node[inner sep=0pt, below of = decoder, xshift=0.75cm, yshift=-0.125cm] (traffic-light)
        {\includegraphics[width=1.4cm]{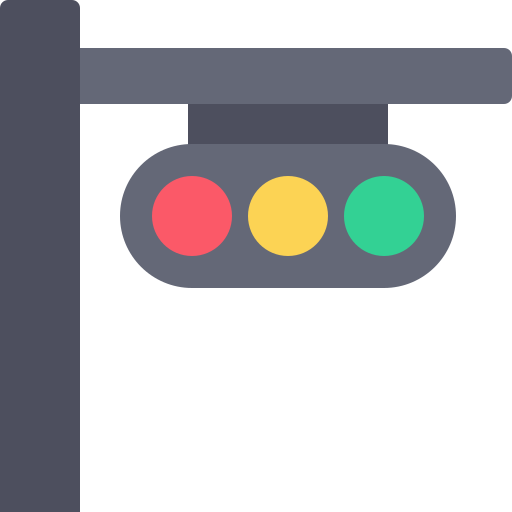}};
        \draw [arrow] (rgb-emb.south) -- +(0, -0.875) -| (decoder.north);
        \draw [arrow] (decoder.south-| stop-sign.north) -- (stop-sign.north);
        \draw [arrow] (decoder.south-| traffic-light.north) -- (traffic-light.north);

        \node (mlp1) [module, above of = cross-semantic-generation-module, align=center, minimum width=1.5cm,yshift=0.625cm, xshift=-1cm] {MLP};
        \node (mlp2) [module, right of = mlp1, align=center, minimum width=1.5cm,yshift=0cm, xshift=0cm] {MLP};
        \node (dash-box2) [dash-box, below of=mlp1, xshift=1cm, yshift=-0.25cm, minimum width=3.5cm, minimum height=2.5cm]{};
        \node (rgb-distribution) [ellipse, right of=dash-box2, xshift=-1.5cm, yshift=-0.25cm, fill=emb-c!60, opacity=0.5, minimum height=1.75cm, minimum width=1.5cm, rotate=-45] {};
        \node (lidar-distribution) [ellipse, left of=dash-box2, xshift=1.5cm, yshift=0cm, fill=emb-c!60, opacity=0.5, minimum height=2cm, minimum width=1.5cm, rotate=60, label={[label distance=0cm, xshift=-1.5cm, yshift=0.5cm, color=white]-90:\small Contrastive Loss}] {};

        \node (shared-emb) [emb, above right of=dash-box2, minimum width=0.75cm, xshift=-0.25cm, yshift=-0.5cm] {{\textcolor{white}{128}}};
        \node[inner sep=0pt, below of = dash-box2, xshift=0.875cm, yshift=-0.5cm] (topdown-seg)
        {\includegraphics[width=1.75cm]{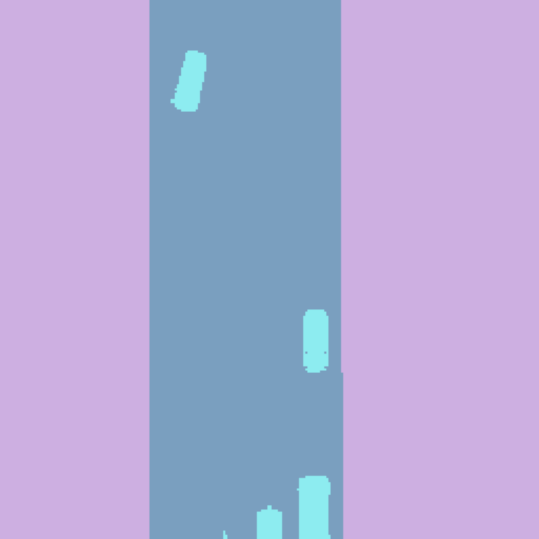}};
        \node[inner sep=0pt, left of = topdown-seg, xshift=-2.25cm] (front-seg)
        {\includegraphics[width=4.5cm]{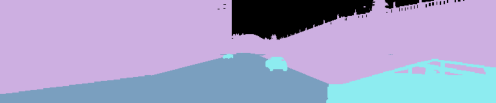}};

        \draw [arrow] (rgb-emb.south) -- +(0, -0.875) -| (mlp1.north);
        \draw [arrow, draw=gray] (lidar-emb.south) -- +(0, -0.875) -| (mlp2.north);
        \draw [arrow] (mlp1.south) to[out=-45,in=130] (rgb-distribution.north west);
        \draw [arrow, draw=gray] (mlp2.south) to[out=-135,in=45] (lidar-distribution.east);
        \draw [arrow] (rgb-distribution) -- (rgb-distribution |- topdown-seg.north) ;
        \draw [arrow, draw=gray] (lidar-distribution) |- (front-seg.east) ;

        \node[inner sep=0pt, above of = penalty-based-imitation-learning-module, xshift=0cm, yshift=3.5cm] (concatenate)
        {\includegraphics[width=0.5cm]{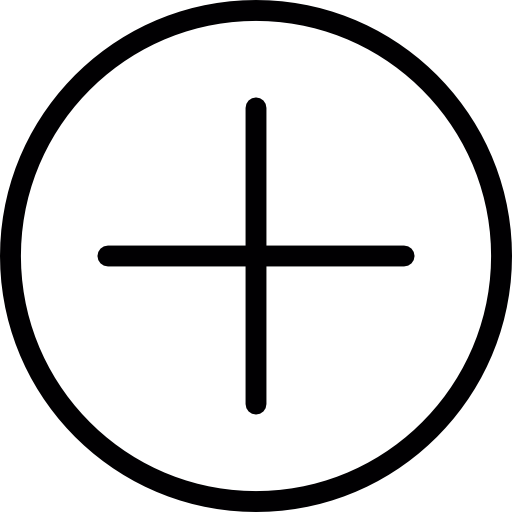}};
        \node (measurements) [emb, right of=concatenate, xshift=-0.5cm, yshift=0cm, minimum width=0.5cm, label = {[label distance=0cm]15:\small Measurements}] {4};
        \node (emb-con) [emb, below of=concatenate, xshift=0cm, yshift=1cm, minimum width=4.5cm] {{\textcolor{white}{1156}}};

        \node (dash-box4) [dash-box, below of=emb-con, draw=black, xshift=2cm, yshift=-3.75cm, minimum width=3.25cm, minimum height=9.5cm]{};
        \node[inner sep=0pt, above of = dash-box4, xshift=0cm, yshift=0.8cm, label={[label distance=0cm]90:\small Red Light Penalty}] (red-light-penalty) {\includegraphics[width=2cm]{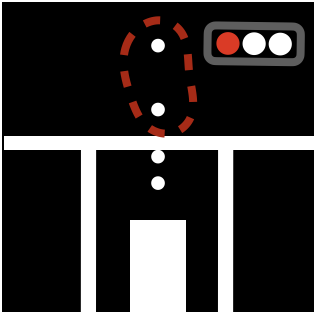}};      

        \node[inner sep=0pt, below of = red-light-penalty, xshift=0cm, yshift=-1cm, label={[label distance=0cm]90:\small Stop Sign Penalty}] (stop-sign-penalty) {\includegraphics[width=2cm]{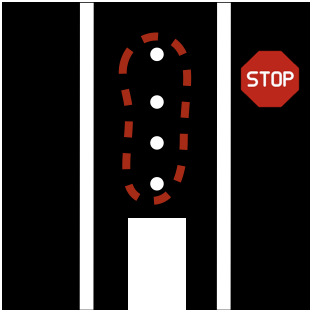}};      

        \node[inner sep=0pt, below of = stop-sign-penalty, xshift=0cm, yshift=-1cm, label={[label distance=0cm]90:\small Speed Penalty}] (speed-penalty) {\includegraphics[width=2cm]{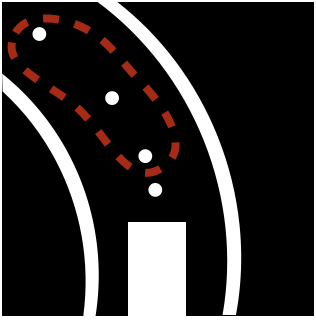}};

        \node (mlp3) [module, below of = emb-con, align=center, minimum width=2.5cm,yshift=0.5cm, xshift=-2.5cm] {MLP};
        \node (emb-64) [emb, below of=mlp3, xshift=0cm, yshift=0.5cm, minimum width=2.5cm] {{\textcolor{white}{64}}};
        \node (gru-decoder) [module, below of = emb-64, align=center, minimum width=2.5cm,yshift=0.5cm] {GRU};
        \node (goal-location) [emb, right of=gru-decoder, xshift=0cm, yshift=0cm, minimum width=0.5cm, minimum height=1cm, label = {[label distance=0cm, align=center, font=\small\linespread{0.9}\selectfont]90: Goal\\ Location}] {};
        \node (waypoints1) [emb, below of=gru-decoder, xshift=-0.125cm, yshift=0.625cm, minimum width=2.25cm, fill=white] {};
        \node (waypoints2) [emb, below of=gru-decoder, xshift=0cm, yshift=0.5cm, minimum width=2.25cm, fill=white] {};
        \node (waypoints3) [emb, below of=gru-decoder, xshift=0.125cm, yshift=0.375cm, minimum width=2.25cm] {{\textcolor{white}{Waypoints}}};

        \node (pid) [module, below of = waypoints2, align=center, minimum width=2.5cm,yshift=0.5cm] {PID};
        \node (dash-box3) [dash-box, below of=pid, draw=black, xshift=0.375cm, yshift=-0.125cm, minimum width=3.5cm, minimum height=1.75cm]{};
        \node[inner sep=0pt, left of = dash-box3, xshift=1.25cm, yshift=0cm] (steer-brake) {\includegraphics[width=1.5cm]{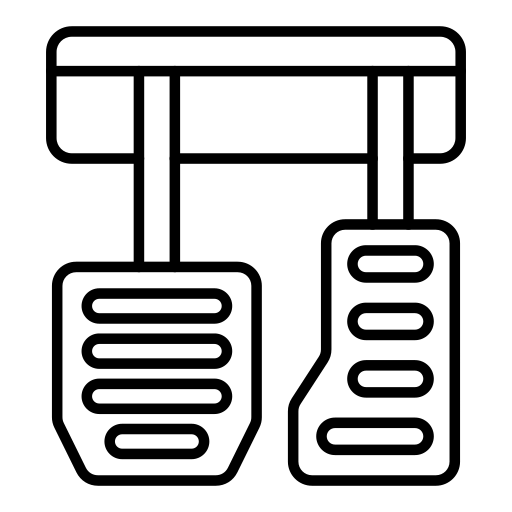}};    
        \node[inner sep=0pt, right of = dash-box3, xshift=-1.25cm, yshift=0cm] (throtte) {\includegraphics[width=1.5cm]{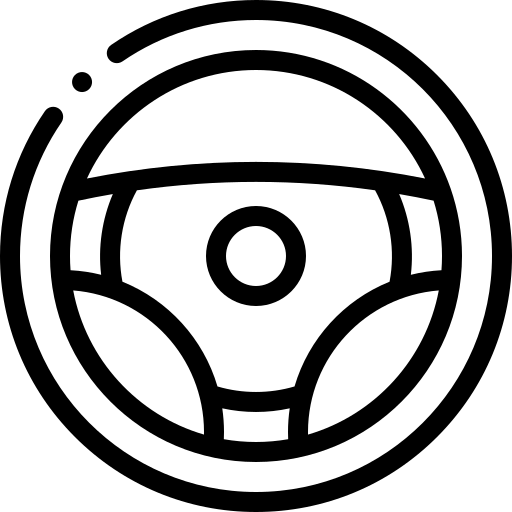}};    
        \draw [dashed, arrow, draw=emb-c!60] (dash-box1.east) -- +(0.75, 0) |- (concatenate.west);
        \draw [dashed, arrow, draw=emb-c!60] (dash-box2.east) -- +(0.75, 0) |- (concatenate.west);

        \draw [arrow, draw=emb-c!60] (measurements.west) -- (concatenate.east);
        \draw [arrow] (concatenate.south) -- (emb-con.north);
        \draw [arrow] (emb-con.south) -- +(0, -0.25) -| (mlp3.north);
        \draw [arrow] (mlp3.south) -- (emb-64.north);
        \draw [arrow] (emb-64.south) -- (gru-decoder.north);
        \draw [arrow] (goal-location.west) -- (gru-decoder.east);
        \draw [arrow] (gru-decoder.south) -- (waypoints2.north);
        \draw [arrow] (gru-decoder.south) to[out=-90,in=-90] +(-1, 0.5) to[out=90,in=135] (gru-decoder.north);
        \draw [arrow] (waypoints3.south-| pid.north) -- ( pid.north);
        \draw [arrow] (pid.south) -- ( pid.north|- dash-box3.north);
        \draw [dashed, arrow] (dash-box4.west |- waypoints3.east) -- ( waypoints3.east) node[above,midway, color=dark-red] {\small Penalties};
        
    \end{tikzpicture}}
    \caption{\small An Overview of Our Penalty-based Imitation Learning with Cross Semantics Generation. 
    }
    \label{fig:Architecture}
\end{figure*}

\subsection{Contrastive-based Cross Semantics Generation}
Our approach's motivation is based on the fact that multi-modality inputs have common information and also their own unique information. For instance, the vehicle's and pedestrian's shape and location are the shared information of LiDAR and camera input. Unique information refers to information that is not present in the other source of modality. Extracting and aligning the common information from both modalities can capture the global context within multi-modalities. In this section, we address two problems. 1) How to find the shared information of these two modalities? 2) How to align the extracted shared information in the same space?

To address the first problem, we propose the \textbf{cross semantics generation} approach. We utilize the LiDAR input to produce semantic segmentation of the camera input and use the camera input to generate semantic segmentation of the LiDAR pseudo image input. As Figure \ref{fig:Architecture} demonstrates, the information flows of the LiDAR (\textcolor{gray}{gray}) and camera (black) semantic information generation are crossed since we used the information from one modality to generate the semantic segmentation of the other modality. This is the reason that we name our approach cross semantics generation. The RGB image and the BEV of the LiDAR pseudo image are fed into two residual networks \cite{DBLP:journals/corr/HeZRS15} to extract the features of these two modalities. We utilize two linear layers to extract the information shared by the camera and LiDAR. 
The feature embeddings are leveraged to generate the semantic segmentation because 1) semantic segmentation has less noise than the original data, and 2) with semantic segmentation, we filter out the unique information, such as traffic light color and traffic sign patterns. In our setup, the semantic segmentation contains 4 channels: the drivable area, the non-drivable area, the objects (vehicles and pedestrians) in the drivable areas, and others. 

We define $y_{\textrm{front}}$ as the ground truth front segmenation tensor with the shape of $160 \times 768 \times 4$ and $\hat{y}_{\textrm{front}}$ as the output of the front view decoder with the same shape. We also define $y_{\textrm{td}}$ as the ground truth top-down segmentation tensor with the shape of $256 \times 256 \times 4$ and $\hat{y}_{\textrm{td}}$ as the output of the top-down view decoder with the same shape. The reconstruction losses can be written as $\mathcal{L}_\textrm{front} = \mathcal{L}_{\textrm{CE}}(\hat{y}_{\textrm{front}},y_\textrm{front})$ and $\mathcal{L}_{\textrm{td}} = \mathcal{L}_{\textrm{CE}}(\hat{y}_{\textrm{td}},y_{\textrm{td}})$, where $\mathcal{L}_{\textrm{CE}}(\cdot)$ indicates the cross entropy loss. $\mathcal{L}_\textrm{front}$ and $\mathcal{L}_{\textrm{td}}$ represent the functions of front reconstruction loss and top-down reconstruction loss, respectively. 

To address the second problem, we leverage the contrastive loss to align the extracted shared information into the same space. The primary goal of contrastive loss is to learn an embedding space where similar examples are placed closer together while dissimilar examples are placed farther apart. This is achieved by defining a distance metric between the embeddings and then minimizing the distance between similar examples while maximizing the distance between dissimilar examples. The embeddings extracted from LiDAR and camera inputs are mapped into two Gaussian distributions by generating mean and variance using MLP networks. Suppose $P_1$ and $P_2$ represent two batches of Gaussian distributions derived from RGB and LiDAR inputs. Each batch contains $N_b$ number of distributions. The contrastive loss can be formulated as follows:
\begin{equation}
    \label{contrastive loss batch}
    \mathcal{L}_{\textrm{align}}({P_1},{P_2}) = mean(\boldsymbol{E} \odot \boldsymbol{D} + (1-\boldsymbol{E}) \odot \max\{0, \epsilon_\textrm{a} - \boldsymbol{D}\})
\end{equation}
where $\bm{D}$ is the matrix consisting of $d_{ij} = d(p_1^i,p_2^j)$. $p_1^i$ is the $i$-th distribution in ${P_1}$ and $p_2^j$ is the $j$-th distribution in ${P_2}$. $d(p_1^i,p_2^j)$ indicates the metric to measure the difference of two distributions. $\boldsymbol{E}$ is the matrix consisting of $e_{ij}$ which indicates if $p_1^i$ and $p_2^j$ belong to the same frame.  $\epsilon_\textrm{a}$ is the threshold difference between similar samples and dissimilar samples. $\odot$ represents the Hadamard product of two matrices. $mean(\cdot)$ represents the mean value of one tensor. 
In our setting, we use the symmetric version of Kullback–Leibler divergence to measure the distance of two distributions:
\begin{equation}
    \label{eq:symmetric kl divergence}
    d(p_1^i, p_2^j) = \frac{1}{2} \textrm{KL}(p_1^i||p_2^j) + \frac{1}{2}\textrm{KL}(p_2^j||p_1^i) \text{,}
\end{equation}
where $\textrm{KL}(\cdot)$ is the KL-divergence metric as the metric to measure the difference between two distributions. The sampled shared embedding is concatenated with image and LiDAR embeddings, along with measurements (speed, throttle, steer, and brake from the previous frame). This combined information serves as the input for the penalty-based imitation learning module.

\subsection{Auxiliary Tasks}
Auxiliary tasks have been proven to be efficient in many learning approaches with two main advantages: 1) Guaranteeing the important information flows remain in the networks. This information flow is critical for the decision-making network. 2) Guiding the direction of gradient descent during training, leading the neural network to a more optimal location in the weight space. In this autonomous driving task, we introduce two extra auxiliary tasks: traffic light classification and stop sign classification.
\subsubsection{Traffic Light Classification}
The output of the traffic light decoder should be a vector of 4, which indicates four states: red light, yellow light, green light, and none in the current frame. We then define $y_{\textrm{light}}$ as the ground truth traffic light vector of length 4 and $\hat{y}_{\textrm{light}}$ as the output of the traffic light decoder with the same shape. We use cross-entropy loss $\mathcal{L}_{\textrm{light}} = \mathcal{L}_{\textrm{CE}}(\hat{y}_{\textrm{light}}, y_\textrm{light})$ for this task.
\subsubsection{Stop Sign Classification}
 The output of the stop sign decoder should have a vector of 1, indicating if a stop sign exists in the current frame. The ground truth stop sign vector of length 1 and the output of the stop sign decoder with the same shape are defined as $y_\textrm{stop}$ and $\hat{y}_\textrm{stop}$, respectively. Also, binary cross-entropy loss $\mathcal{L}_{\textrm{stop}} = \mathcal{L}_{\textrm{BCE}}(\hat{y}_\textrm{stop}, y_\textrm{stop})$ is used for this task 

\vspace{5mm}
Note that these two tasks are trained simultaneously with policy generation networks and cross semantics generation tasks. In summary, the final loss function can be calculated as:
\begin{equation}
\begin{aligned}
        \mathcal{L}_\textrm{final} = & \mathcal{L}_{\textrm{pl}} + \eta_1 \mathcal{L}_\textrm{front} + 
        \eta_2 \mathcal{L}_{\textrm{td}} + \eta_3 \mathcal{L}_\textrm{light} +  
        \eta_4 \mathcal{L}_\textrm{stop} + \eta_5 \mathcal{L}_\textrm{align} \text{,}
\end{aligned}
\label{eq:new-designed objective}
\end{equation}
where $\mathcal{L}_{\textrm{pl}}$ is the policy learning loss defined in Equation \eqref{eq:policy learning loss}; $\mathcal{L}_\textrm{front}$ and $\mathcal{L}_{\textrm{td}}$ are front view and top-down view segmentation losses; $\mathcal{L}_\textrm{align}$ is the contractive alignment loss defined in Equation \eqref{contrastive loss batch}; $\mathcal{L}_\textrm{light}$ and $\mathcal{L}_\textrm{stop}$ are losses for auxiliary tasks, namely traffic light classification and stop sign classification. $\eta_1$,  $\eta_2$, $\eta_3$, $\eta_4$, $\eta_5$ are weights to balance these losses.

 \subsection{Penalty-based Imitation Learning with Constraint Optimization}
Our investigation revealed that the objective function used in imitation learning and the metric used for evaluating autonomous driving performance are not consistent. Hence, achieving a low loss in the objective function does not necessarily guarantee a high driving score and route completion. We identified two possible factors contributing to this discrepancy.
 
\begin{itemize}
    \item The expert agent still makes mistakes when generating the dataset. Sometimes, the expert agent runs a red light and violates the stop sign rule.
    \item The objective function is not sensitive to serious violations of the traffic rules, i.e., the violation of red lights and stop signs. The average objective function loss may not increase too much when violating the traffic rules, although this violation may cause serious consequences, which result in a huge drop in driving score and route completion.
\end{itemize}

We aim to constrain the objective function of imitation learning to incorporate traffic rules. 
Traffic rules can be represented as constraint functions that specify the conditions the optimization problem must satisfy. Our approach focuses on three specific aspects: running red lights, ignoring stop signs, and failing to slow down when turning. These are the primary issues we observed with our vanilla imitation learning method. To address these issues, we propose three corresponding penalties that can be used to quantify and penalize these violations.

\subsubsection{Red Light Penalty}
For the red light violation, we design a red light penalty as follows:
\begin{equation}
    {\mathcal{P}}_{\rm tl} = \mathbb{1}_{\rm red}\cdot\sum_{t=1}^{T}c_i\cdot {\rm max}\{0, \hat{y}_t - y_{\textrm{stop}}\}  \text{,}
\end{equation}
where $\hat{y}_t$ represents the $y$ axis value in the $t$-th predicted waypoints $\bm{\hat{w}_t} = (\hat{x}_t,\hat{y}_t)$ and $y_{\textrm{stop}}$ denotes the distance between the ego car and the stop line at the intersection along the forward direction of the ego car ($y$ axis). The weight parameter is denoted by $c_i$, and the sum of all weight parameters is equal to one. $\mathbb{1}_{\rm red}$ indicates the presence of a red light that may affect the agent in the current frame. 

When facing red lights, a red light penalty is added based on the distances of the predicted waypoints beyond the stop line at the intersection. If the predicted waypoints fall within the stop line, the penalty remains at zero. Conversely, if the predicted waypoints exceed the stop line, the total distance between those waypoints and the stop line is computed as the red light penalty. The necessary information for calculating the red light penalty, such as traffic light information and stop line location, is pre-processed and stored in each frame of our dataset.

\subsubsection{Stop Sign Penalty}
Similar to the red light penalty, a stop sign penalty is given when the predicted waypoints violate the stop sign rule. The penalty is formalized as follows:
\begin{equation}
    {\mathcal{P}}_{\rm ss}=\mathbb{1}_{\rm stopsign}\cdot  {\rm max}\{v - \epsilon_v, 0\}
\end{equation}
where $v$ is the estimated speed calculated by 
\begin{equation}
    v = \frac{||\bm{\hat{w}_{0}} - \bm{\hat{w}_{1}}||_2}{\Delta t}
    \label{eq: desired speed}
\end{equation}
The variables $\bm{\hat{w}_0}$ and $\bm{\hat{w}_1}$ represent the first and second predicted waypoints, respectively, while $\Delta t$ indicates the time interval between each frame. The function $\mathbb{1}_{\rm stopsign}$ serves as an indicator for stop sign checking. The maximum speed to pass stop sign tests is denoted by $\epsilon_v$. 

An upper-speed limit $\epsilon_v$ (close to zero) is established for an area affected by a stop sign, and only speeds lower than this limit are permitted for the agent to pass through. If the agent exceeds this speed limit, a penalty is imposed based on its speed. As the training network only generates predicted waypoints, the speed estimated from the waypoints is used to compute the stop sign penalty.

\subsubsection{Curvature Speed Penalty}
A penalty will be enforced if the agent attempts to turn at excessive speed. The rationale behind this penalty is based on human driving experience, as it is commonly known that turning at high speeds can increase the risk of collisions with pedestrians or other objects due to longer braking distance. The speed penalty is defined as follows:
\begin{equation}
    {\mathcal{P}}_{\rm sp}={\rm \sin}(\Delta\delta) \cdot {\max} \{v - v_{\textrm{lb}}, 0\}
\end{equation}
 where $\Delta\delta$ denotes the deviation in direction between the current frame and the next frame. As with the stop sign penalty, the desired speed $v$ is defined by Equation \ref{eq: desired speed}, and $v_{\rm lb}$ represents the lower speed limit. Any speed below this limit is not subject to the speed penalty.

\subsubsection{Objctive Function}
We trained the network to minimize the difference between the predicted and ground truth waypoints, applying an $l_1$-norm-based loss function. Also, cross semantics generation and auxiliary tasks are trained together with the policy generation task. Equation \eqref{eq:new-designed objective} defines the final loss function. By applying the penalties, we formalize the constrained optimization:

By applying the penalties, we formalize the constrained optimization:
\begin{equation}
\label{eq:constraint objective function}
\begin{aligned}
\min_{\boldsymbol{\Theta}} \quad & \mathcal{L}_\textrm{final} \\
\textrm{s.t.} \quad & \mathcal{P}_{\rm tl}, \mathcal{P}_{\rm ss},\mathcal{P}_{\rm sp} = 0\\
\end{aligned}
\end{equation}
where $\mathcal{L}_\textrm{final}$ is the loss function defined in Equation \eqref{eq:new-designed objective}. $\mathbf{\Theta}$ denotes a vector that includes all learnable parameters in the model. It is constrained that the penalties should equal 0 since 
 the predicted waypoints should not violate the traffic rules. 

The Lagrange multiplier strategy can be applied here. We introduce three Lagrange Multiplier $\lambda_1$, $\lambda_2$, $\lambda_3$ and the Lagrange function is defined by:
\begin{equation}
\begin{aligned}
\min_{\boldsymbol{\Theta}} \quad \mathcal{L}_\textrm{final} + \lambda_1 \mathcal{P}_{\rm tl}+ \lambda_2 \mathcal{P}_{\rm ss} +\lambda_3 \mathcal{P}_{\rm sp}
\end{aligned}
\end{equation}

This is the final objective function to optimize. For simplicity, these Lagrange multipliers $\lambda_1$, $\lambda_2$, $\lambda_3$ are considered fixed hyper-parameters. Well-chosen $\lambda_1$, $\lambda_2$, $\lambda_3$ are important for optimization. According to our experiments, hyperparameters that are too large influence behaviors in other scenarios while hyperparameters that are too small are not powerful enough for the agent to obey the corresponding traffic rules. The right part of Figure \ref{fig:Architecture} demonstrates the process of penalty-based imitation learning in detail. 

\vspace{-5mm}
\section{Experiments}
\label{sec:experiments}
\subsection{Training Dataset}
Obtaining realistic driving data is challenging. Therefore, we opted to use the CARLA simulator to gather training data that has been processed by the expert policy. Our training dataset comprises approximately 2500 routes through junctions in 8 different towns. These routes have an average length of 100 meters and around 1000 routes along curved highways with an average length of 400 meters. The expert policy we use is the updated one in TransFuser+. For each frame, We collect 
\begin{itemize}
    \item The forward, left 60 degrees, and right 60 degrees images with the resolutions of $400 \times 300$ and their corresponding semantic segmentations. 
    \item The cloud point of LiDAR with 180 degrees in front of the vehicle and the top-down semantic segmentation.
    \item The steer, throttle, brake action values, and speed measurements in the current frame. 
    \item The traffic light information, which includes the stop line position and traffic light status.
    \item The stop sign information indicates if the current frame is influenced by a stop sign. 
    \item The future waypoints in the next 4 frames. 
\end{itemize}
Note that the interval for each frame is 0.5 seconds, indicating 2FPS frequency.  

\subsection{Metrics}
Route completion and infraction scores are used to evaluate the agent's behavior.

\noindent\textbf{Route Completion}
Route completion (RC) refers to the proportion of the completed route out of the whole route. Suppose $R_j$ means the route completion proportion in route $j$ and $N_\textrm{route}$ means the number of routes. The RC can be defined by:
\begin{equation}
    RC = \frac{1}{N_\textrm{route}} \sum_{j}^{N_\textrm{route}} R_j
\end{equation}

\noindent\textbf{Infraction Score}
Infraction Score (IS) is used to measure the driving behavior of the agent. We define $p_k$ as the penalty for an infraction instance, $k$ is incurred by the agent, and $n_k$ is the number of occurrences of infraction instance $k$. Then, the infraction score can be defined by:
\begin{equation}
    \label{eq:infraction score}
    IS = \prod_{k}^{\{\rm Ped, Veh, Stat, Red, Stop\}} s_k^{n_k}\text{.}
\end{equation}
The infraction instances include collision with a pedestrian, collision with a vehicle, collision with static layout, red light violations, and stop sign violations. The penalty scores $s_k$ for them are 0.5, 0.60, 0.65, 0.7, and 0.8, respectively. $n_k$ is the number of infractions. 

\noindent\textbf{Driving Score}
Driving Score (DS) aims to measure the overall driving performance. It is defined by the weighted average of the route completion with an infraction multiplier:
\begin{equation}
    DS = \frac{1}{N_\textrm{route}} \sum_{j} ^{N_\textrm{route}} R_j IS_j \text{,}
\end{equation}
where $R_j$ and $IS_j$ are route completion, infraction score $IS$ \eqref{eq:infraction score} for $j$-th route.


\subsection{Test Results}
We use CARLA Leaderboard - Town05 Long Benchmark and Longest6 Benchmark to evaluate our model. Town05 Long Benchmark contains 10 routes, all of which are over 2.5 km. This benchmark is also used by InterFuser, and TransFuser. Longest6 benchmark is proposed in TransFuser+ with increased traffic density, and challenging pre-crash traffic scenarios.

\begin{table*}
\vspace{-10pt}
\centering
\begin{minipage}[t]{0.47\linewidth}
\centering
\resizebox{\linewidth}{!}{%
\begin{tabular}{l c c c}
\toprule
\multirow{2}{*}{Model} & \makecell[c]{Driving\\score} & \makecell[c]{Route\\compl.} & \makecell[c]{Infrac.\\score} \\
\cmidrule{2-4}
& $\%, \uparrow$ & $\%, \uparrow$ & $\%, \uparrow$ \\
\midrule
P-CSG(Ours) & $\textbf{59.10} \pm 4.07$ & $86.84 \pm 2.99$ & $\textbf{0.68} \pm 0.06$ \\
TransFuser \cite{Prakash2021CVPR} & $34.50 \pm 2.54$ & $61.16 \pm 4.75$ & $0.56 \pm 0.06$ \\
TransFuser+ \cite{Chitta2022PAMI} & $36.19 \pm 0.90$ & $70.13 \pm 6.80$ & $0.51 \pm 0.03$ \\
InterFuser \cite{shao2022interfuser} & $50.64 \pm 3.51$ & $89.13 \pm 4.12$ & $0.57 \pm 0.05$ \\
LAV \cite{chen2022learning} & $45.20 \pm 6.36$ & $\textbf{91.55} \pm 5.61$ & $0.49 \pm 0.06$ \\ 
\bottomrule
\end{tabular}
}
\caption{\footnotesize CARLA Leaderboard - Town 05 Long Benchmark.}

\label{tab:town_05_long_benchmark}
\end{minipage}%
\hspace{0.04\linewidth}
\begin{minipage}[t]{0.47\linewidth}
\centering
\resizebox{\linewidth}{!}{%
\begin{tabular}{l c c c}
\toprule
\multirow{2}{*}{Model} & \makecell[c]{Driving\\score} & \makecell[c]{Route\\compl.} & \makecell[c]{Infrac.\\score} \\
\cmidrule{2-4}
& $\%, \uparrow$ & $\%, \uparrow$ & $\%, \uparrow$ \\
\midrule
P-CSG(Ours) & $\textbf{49.21} \pm 2.93$ & $ 86.24 \pm 2.46$ & $\textbf{0.56} \pm 0.04$ \\
WOR \cite{chen2021learning} & $20.53 \pm 3.12$ & $48.47 \pm 3.86$ & $0.56 \pm 0.03$ \\
LAV \cite{chen2022learning} & $32.74 \pm 1.45$ & $70.36 \pm 3.14$ & $0.51 \pm 0.02$ \\
Late Fusion \cite{sobh2018end} & $22.47 \pm 3.71$ & $83.30 \pm 3.04$ & $0.27 \pm 0.04$ \\
TransFuser+ \cite{Chitta2022PAMI} & $47.30 \pm 5.72$ & $\textbf{93.38} \pm 1.20$ & $0.50 \pm 0.06$ \\
\bottomrule
\end{tabular}
}
\caption{\footnotesize CARLA Leaderboard - Longest6 Benchmark. }
\label{tab:longest6_benchmark}
\end{minipage}
\vspace{-25pt}
\end{table*}



The evaluation results in Table \ref{tab:town_05_long_benchmark} indicate that our model significantly increases driving scores and infraction penalties compared to other baselines. Table \ref{tab:longest6_benchmark} demonstrates the performance of our model in the Longest6 Benchmark provided by Transfuser+. Our model surpasses all other baselines in driving and infraction scores. While our model may not attain the top route completion score, its highest infraction score and driving score suggest that it embraces a more cautious approach to mitigating traffic rule violations, ultimately leading to the highest driving score.

\subsection{Ablation Studies}
In this subsection, we analyze the influences of different penalty weights for corresponding traffic rules. The ablation study focuses on significant hyper-parameters, namely $\lambda_1$, $\lambda_2$, and $\lambda_3$, which correspond to the penalty weights associated with the red light, speed, and stop sign. Our optimal model employs default weights of $\lambda_1 = 0.5$, $\lambda_2 = 0.05$, and $\lambda_3 = 0.5$. As Table \ref{tab:ablation} demonstrates,
\begin{wraptable}{r}{0.45\textwidth}
   \vspace{2pt}
    \centering
    \resizebox{0.45\textwidth}{!}{%
    \begin{tabular}{l c c c}
    \toprule
    \multirow{2}{*}{Model} & \makecell[c]{Driving\\score} & \makecell[c]{Route\\compl.} & \makecell[c]{Infrac.\\score} \\
    \cmidrule{2-4}
    & $\%, \uparrow$ & $\%, \uparrow$ & $\%, \uparrow$ \\
    \midrule
    P-CSG(Ours) & $ \textbf{59.10} \pm 4.07$ & $ 86.84 \pm 2.99$ & $\textbf{0.68} \pm 0.06$ \\
    No CSG & $45.67 \pm 5.51$ & $84.82 \pm 2.25$ & $ 0.52 \pm 0.10$ \\  
    No Penalty & $37.32 \pm 3.58$ & $78.41 \pm 11.32$ & $0.54 \pm 0.21$ \\
    $\lambda_1 = 0.7$ & $54.36 \pm 1.78$ & $84.37 \pm 5.01$ & $0.64 \pm 0.04$ \\
    $\lambda_1 = 0.3$ & $56.80 \pm 2.22$ & $89.60 \pm 1.90$ & $0.60 \pm 0.02$ \\
    $\lambda_2 = 0.07$ & $51.54 \pm 1.50$ & $81.49 \pm 1.42$ & $0.63 \pm 0.02$ \\
    $\lambda_2 = 0.03$ & $54.59 \pm 2.88$ & $88.12 \pm 1.95$ & $0.64 \pm 0.02$ \\
    $\lambda_3 = 0.7$ & $55.98 \pm 3.00$ & $\textbf{89.81} \pm 3.33$ & $0.63 \pm 0.02$ \\
    $\lambda_3 = 0.3$ & $51.98 \pm 1.93$ & $84.66 \pm 3.53$ & $0.63 \pm 0.01$ \\
    \bottomrule
    \end{tabular}
    }
    \caption{Ablation Study.}
    \label{tab:ablation}
    \vspace{-20pt}
\end{wraptable}
two extra weights for each penalty are selected for comparison. We also provide the results of models without CSG and penalties for comprehensive analysis. 
The results of different penalty weights are also listed in the table. We found that assigning greater weight to more severe violations will increase the performance of our model. For instance, we apply greater penalties for the red light and the stop sign violations than overspeeding by turning since those two violations cause more serious consequences.

\section{Robustness Study}
\label{sec:robustness study}

\begin{figure}[ht]
    \centering
    \begin{adjustbox}{width=0.8\textwidth}
    \begin{tikzpicture}[node distance=2cm]
        \node[inner sep=0pt, label={[label distance=0cm]180:\small (a)}] (figure1){\includegraphics[width=8cm]{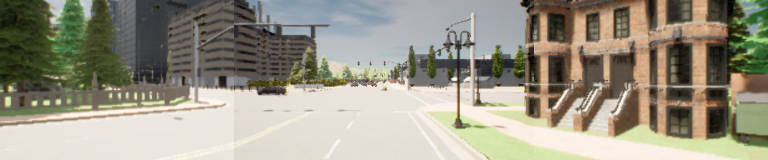}};
        \node[inner sep=0pt, below of = figure1, label={[label distance=0cm]180:\small (b)}] (figure2){\includegraphics[width=8cm]{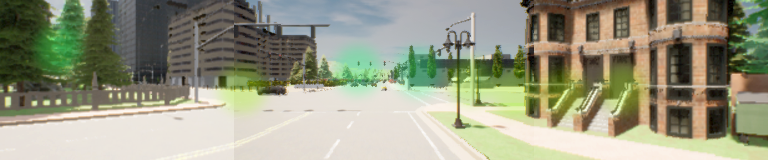}};
        \node[inner sep=0pt, below of = figure2, label={[label distance=0cm]180:\small (c)}] (figure3){\includegraphics[width=8cm]{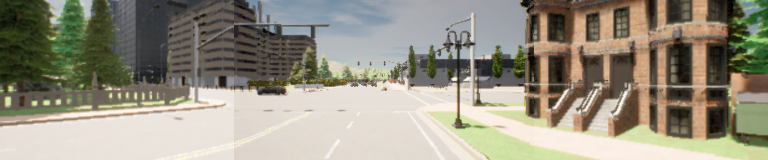}};
    \end{tikzpicture}
    \end{adjustbox}
    \caption{\textbf{Qualitative Attack Results on P-CSG.} (a) Original RGB input, (b) Dot Attack with nine trained dots, (c) FGSM Attack with \(\epsilon = 0.01\). The subtle FGSM perturbation in (c) is hard to spot compared to (a), showcasing the method's ability to create changes that are invisible to the human eye but still mislead models.}
    \label{fig:attack}
\end{figure}
Given the life-critical nature of autonomous driving, examining the robustness of neural network models against such vulnerabilities becomes essential. In this regard, we conduct a comprehensive comparative analysis to evaluate the robustness of our model. We benchmark its performance against two state-of-the-art methods — Transfuser+ and Interfuser. We focus on two specific types of white-box sensor attacks. These are \textbf{Fast Gradient Sign Method (FGSM) Attack} and \textbf{Dot Attack}.
\footnote{Interfuser* has the same PID Controller as P-CSG and Transfuser+ to ensure a fair comparison.}

\begin{itemize}
    \item \textbf{FGSM Attack} is notable for its computational efficiency and minimal perceptibility to the human eye. It uses the gradient of the model's loss function to slightly alter the input image, creating an adversarial image that looks almost identical but can mislead the model.
    \item \textbf{Dot Attack} uses special stickers with dot patterns on camera lenses to subtly blur parts of the image. These near-invisible dots mislead deep learning models without being noticeable to humans. Each dot's color blends the original and a new RGB value, decreasing transparency from the center outward. This technique effectively disrupts model performance with minimal visual impact. Unlike computational attacks like FGSM, Dot Attacks are easily deployed in the real world, requiring only the application of a sticker to the camera lens.

\end{itemize}

\subsection{Results}

In our evaluation, the \textbf{infraction score} serves as the primary metric, encapsulating various collision scenarios and traffic violations to comprehensively assess safety. 
As shown in Table \ref{tab:attack_results}, P-CSG demonstrates superior performance in two distinct adversarial scenarios: high-intensity, per-frame FGSM Attack, and uniformly applied, low-intensity Dot Attack.

\begin{wraptable}{r}{0.55\textwidth}
    \vspace{-20pt}
    \centering
    \renewcommand{\arraystretch}{} 
    \resizebox{0.55\textwidth}{!}{%
    \begin{tabular}{llccc}
    \toprule
    \multirow{2}{*}{Attack} & \multirow{2}{*}{Model} & \makecell[c]{Driving\\score} & \makecell[c]{Route\\compl.}& \makecell[c]{Infrac.\\score} \\
    \cmidrule{3-5}
     & & $\%, \uparrow$ & $\%, \uparrow$ & $\%, \uparrow$ \\
    \midrule
    \multirow{4}{*}{\makecell[c]{FGSM Attack\\$(\epsilon = 0.01)$}} & P-CSG(Ours) & $ \mathbf{12.80} \pm 1.35$ & $ 21.70 \pm 2.87$ & $\mathbf{0.59} \pm 0.02$ \\
    & Transfuser+ & $2.69 \pm 0.73$ & $42.44 \pm 3.65$ & $0.22 \pm 0.06$ \\
    & Interfuser* & $8.37 \pm 1.06$ & $\mathbf{72.04} \pm 4.15$ & $0.15 \pm 0.04$ \\
    \midrule
    \multirow{3}{*}{Dot Attack} & P-CSG(Ours) & $ \mathbf{36.60} \pm 3.17$ & $\mathbf{75.53} \pm 6.43$ & $\mathbf{0.56} \pm 0.07$ \\
    & Transfuser+ & $0.53 \pm 0.21$ & $54.06 \pm 5.48$ & $0.02 \pm 0.01$ \\
    & Interfuser* & $12.56 \pm 2.15$ & $73.62 \pm 7.32$ & $0.19 \pm 0.02$ \\
    \bottomrule
    \end{tabular}
    }
    \caption{Attack Results.}
    \label{tab:attack_results}
    \vspace{-30pt}
\end{wraptable}
Specifically, P-CSG achieved an infraction score that was at least \(2.5\) times higher than that of competing models in both attack scenarios tested. Under the low-intensity Dot Attack, P-CSG not only maintained high safety standards, indicated by a higher infraction score, but also completed more routes, as reflected in a higher route completion metric. Conversely, in scenarios involving high-intensity FGSM Attacks, P-CSG implemented a more conservative strategy aimed at enhancing security. Its higher infraction score shows this, even though it led to fewer completed routes.
Most importantly, P-CSG consistently reaches the highest cumulative driving score in both types of attacks, confirming its robustness and adaptability in different challenging conditions. 


\section{Conclusion}
\label{sec:conclusion}
In this paper, we improve the multi-modality fusion technologies and policy learning methods based on penalties for autonomous driving. 
We observe that our contrastive learning-based multi-modal fusion method is helpful in extracting the global content of various modalities. 
Furthermore, we found that incorporating penalties based on traffic rules into an imitation learning pipeline can enhance the agent's adherence to those traffic rules.
We also compare our proposed model with other baselines under Dot Attacks and FGSM Attacks. The results show that our approach is more robust against these attacks. 
We aspire for our proposed penalty-based imitation learning approach to introduce a fresh perspective into the domain of end-to-end autonomous driving to enhance autonomous agents' compliance with traffic rules. 

One of our main objectives for future works is to include more types of penalties in the learning pipeline. These penalties could be designed based on multiple other traffic rules like speed limit, lane changing, and following distance.
Scaling our method with a larger neural network and dataset is another direction of investigation, aimed at determining whether a larger model and data can effectively enhance overall performance.

\newpage
\bibliographystyle{splncs04}
\bibliography{main}

\end{document}